\documentclass[letterpaper, 10 pt, conference]{ieeeconf}  %
\IEEEoverridecommandlockouts                              %
\usepackage[T1]{fontenc} 
\usepackage{enumerate}
\usepackage{amsfonts}
\usepackage[colorinlistoftodos]{todonotes}
\usepackage[font=small,skip=12pt]{caption}
\usepackage{float}
\usepackage{subcaption}
\usepackage[ruled,vlined]{algorithm2e}
\usepackage{algorithmicx}
\usepackage{booktabs}
\makeatletter
\let\NAT@parse\undefined
\makeatother
\usepackage{hyperref}
\usepackage{verbatim}
\usepackage{amsmath}

\algnewcommand{\LeftComment}[1]{\Statex \(\triangleright\) #1}

\title{\LARGE \bf A 2D Surgical Simulation Framework for Tool-Tissue Interaction
}
\author{Yunhai Han$^1$, Fei Liu$^1$, and Michael C. Yip$^1$ \IEEEmembership{Member, IEEE}
\thanks{$^1$Yunhai Han, Fei Liu and Michael C. Yip are with the Advanced Robotics and Controls Lab, University of California San Diego, La Jolla, CA 92093 USA. {\tt\small \{y8han, f4liu, yip\}@ucsd.edu}}%
}
\begin{document}
\maketitle
\thispagestyle{empty}
\pagestyle{empty}
\begin{abstract}
The control and task automation of robotic surgical system is very challenging, especially in soft tissue manipulation, due to the unpredictable deformations.
Thus, an accurate simulator of soft tissues with the ability of interacting with robot manipulators is necessary. In this work, we propose a novel 2D simulation framework for tool-tissue interaction. This framework continuously tracks the motion of manipulator and simulates the tissue deformation in presence of collision detection.
The deformation energy can be computed for the control and planning task.
\end{abstract}

\vspace{-0.3cm}
\section{INTRODUCTION}
\vspace{-0.2cm}
Recently, the expectations placed on surgical robotics have risen sharply. However, currently, surgeons are still required to manipulate the central control unit. In order to reduce surgeon fatigue, recent research has been extended to develop new control algorithms for surgical task automation \cite{Yip2017Automation}. The task can be divided into two parts:
\begin{itemize}
    \item Prediction of tissue deformation after a certain operation (e.g. input control commands).
    \item Generation of operations based on the current and target deformation.
\end{itemize}
This work focuses on the first part, that is to simulate the 2D tissue deformation under the operation of a robotics manipulator.

The rest of the paper is organized as follows: Section \ref{section: methodology} presents the detail of the framework; Section \ref{section: experiments} shows the simulation results; Section \ref{sec:conclusion} gives the conclusion.
\vspace{-0.15cm}
\section{METHODOLOGY} \label{section: methodology}
\vspace{-0.15cm}
The simulation framework can be divided into four modules:
\begin{itemize}
    \item Mesh generation using 2D tissue images
    \item Position-based dynamics methods for tissue simulation
    \item Collision detection method for tool-tissue interaction
    \item Implicit Euler energy computation
\end{itemize}
\vspace{-0.3cm}
\subsection{Mesh Generation}
\begin{figure}[htbp]
\centering
\begin{minipage}[t]{0.22\textwidth}
\centering
\includegraphics[height = 2cm, width = 2cm]{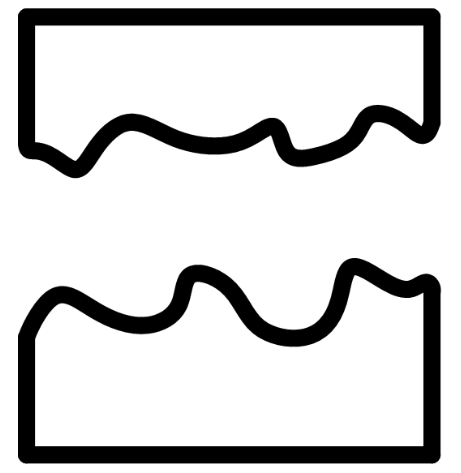}
\end{minipage}%
\begin{minipage}[t]{0.22\textwidth}
\centering
\includegraphics[height = 2cm, width = 2.5cm]{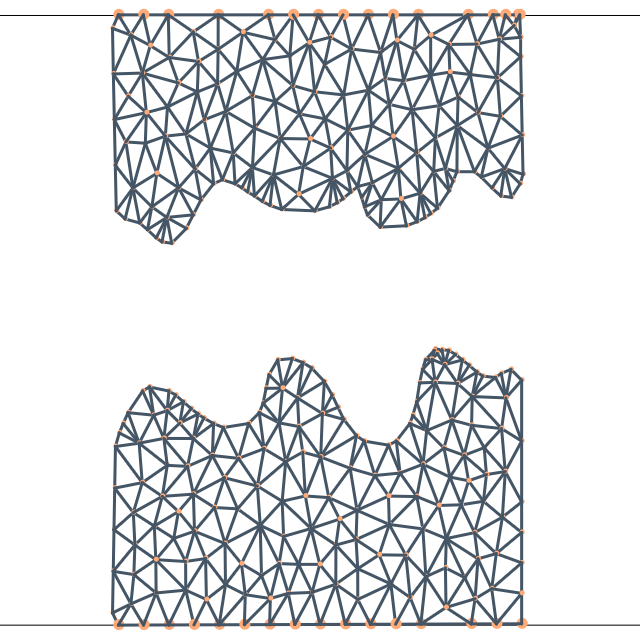}
\end{minipage}%
\centering
\caption{A demonstration of mesh generation. The left image is the tissue image and the right image is the generated triangle meshes.}
\label{fig:mesh_demo}
\end{figure}

Images only describe the geometric properties (e.g. shape) of the tissues, which are incomplete for any physical-based simulations. Hence, it is required to build the triangle/volumetric representations for the simulation of 2D/3D objects from the tissue images.
In this work, a 2D surgical environment is generated using a standard tessellation algorithm (Delaunay Triangulator), as shown Fig. \ref{fig:mesh_demo}.
\subsection{PBD Simulation}
In the recent decades, a position-based dynamics (PBD) method has earned increasing attention and has been shown to provide real-time performance  and can be implemented efficiently \cite{Mull2017PBD}, compared to traditional FEM, such as SOFA framework \cite{Faure_2012}. 


\subsubsection{Simulation Process}
The deformed object is defined as a set of $N$ particles and $M$ constraints. The simulation process is described in Algorithm \ref{alg_PBD}.
\vspace{-0.4cm}
\begin{algorithm}[h]
\SetAlgoLined

$\mathbf{x}^* = \mathbf{x}^{t} + \Delta t \mathbf{v}^t + \Delta t^2 \mathbf{M}^{-1} \mathbf{f}_{ext}(\mathbf{x^t}) $ \Comment{prediction step}
\\


\While{iter $<$ SolverIterations}{
\For{constraint $C \in M$}{
\text{Compute} $\Delta \mathbf{x}$ \Comment{constraint solving step} \\
$\mathbf{x}=\mathbf{x^{*}}+\Delta \mathbf{x}$
}
}
$\mathbf{x}^{t+1} = \mathbf{x}$ \Comment{update position} \\
$\mathbf{v}^{t+1} = \left(\mathbf{x}^{t+1}-\mathbf{x}^{t}\right)/\Delta t$ \Comment{update velocity} \\
\caption{Simulation Process}
\label{alg_PBD}
\end{algorithm} 
\vspace{-0.5cm}

In Algorithm \ref{alg_PBD}, $\mathbf{f}_{ext}$ includes the gravity force and the driving force exerted by the manipulator when the collision is detected. However, in most cases, the driving force is unknown. In this work, instead of applying the driving force on the particles, we directly update their positions. You can image that there is an "invisible" force that leads to the position update $x^{\text{update}}$ in Equ \ref{equ:update}. 
\subsubsection{The Gauss-Seidel Method}
In Algorithm \ref{alg_PBD}, the position correction $\Delta \mathbf{x}$ in each iteration can be computed through Gauss-Siedel Method \cite{Mull2017PBD}.  In this work, the type of constraints are selected as follow:
\begin{itemize}
    \item \textbf{Distance constraint}: The distance constraint between each set of connected particles ($\mathbf{x}_{1}$ and $\mathbf{x}_{2}$) can be satisfied by introducing:
    \begin{equation}
    C_{\text{spring}}\left(\mathbf{x}_{1}, \mathbf{x}_{2}\right)=\left|\mathbf{x}_{1} - \mathbf{x}_{2}\right|-d_0.
    \end{equation}
where $d_0$ is the initial distance indicated by rest spring length.

    \item \textbf{Area conservation}: The area of a triangle, represented by three particles $\mathbf{x}_{1}, \mathbf{x}_{2}$ and $\mathbf{x}_{3}$, can be kept constant by introducing :
    \begin{equation}
C_{\text{area}} \left(\mathbf{x}_{1}, \mathbf{x}_{2}, \mathbf{x}_{3}\right)=\frac{1}{2}\left|\left( \mathbf{x}_{2} - \mathbf{x}_{1} \right) \times \left( \mathbf{x}_{3} - \mathbf{x}_{1} \right)\right|-A_{0}
\end{equation}
where $A_0$ is the initial area of the triangle.
    
\end{itemize}
\vspace{-0.3cm}
\subsection{Collision Detection for Tool-Tissue Interaction}
Using flexible-collision-library\footnote{The usage documents can be found on github: \href{https://github.com/BerkeleyAutomation/python-fcl}{python-fcl}.}, the collision point is defined as the particle $q$ on the mesh when the distance ($d_{*}$) between it and the manipulator is smaller than a fixed threshold $\alpha$. The position update of particle $i$ is:
\begin{equation} \label{equ:update}
x_i^{\text{update}} = 
\left\{
\begin{array}{rl}
    \frac{r - \left\|\mathbf{x}_q-\mathbf{x}_i\right\|}{r}d_{*}\frac{\mathbf{v}_{\text{mani}}}{\left\|\mathbf{v}_{\text{mani}}\right\|},& \text{if} \, r - \left\|\mathbf{x}_q-\mathbf{x}_i\right\| \geq 0 \\
    0,& \text{otherwise}
\end{array}
\right.
\end{equation}
where, $r$ is the circle radius and $\mathbf{v}_{\text{mani}}$ is the velocity of the approaching manipulator.
\subsection{Implicit Euler Energy Computation}
The definition of implicit Euler energy \cite{Mull2017PBD} contains both the inertial and potential energy, i.e.,:
\begin{equation}
E(x)=\frac{1}{2} \left\|\mathbf{x}^{t+1}-\mathbf{x}^{*}\right\|^{2}_\mathbf{M}+\Delta t^{2} E_p(\mathbf{x}^{t+1})
\end{equation}
where, $\mathbf{x}^{*}$ and $\mathbf{x}^{t+1}$ are the particles positions before and after constraint solving step in Algorithm \ref{alg_PBD}, $\Delta t^{2}$ is the simulation time step and $E_p(\mathbf{x}^{t+1})$ is the internal potential energy during constraint solving step, which is the sum of the following two parts : spring elastic energy and area conservation energy.

We make extensive use of the compliance form of elasticity as in \cite{Mull2017PBD}. Thus, the area conservation energy is defined by a quadratic potential energy in terms of the constraint function as, 
\begin{equation}
E_{i}= \frac{1}{2} \mathbf{C}_{i}^T(\mathbf{x}) \mathbf{K}_i \mathbf{C}_{i}(\mathbf{x}) 
\end{equation}
where $\mathbf{C}_{i}=\left[C_{i}^1(\mathbf{x}), C_{i}^2(\mathbf{x}), \cdots, C_{i}^n(\mathbf{x})\right]^T$, where $j \in \lbrace \text{spring}, \text{area} \rbrace $, and $n \in \lbrace \text{s}, \text{a} \rbrace $ represents that $s$ the number of springs and $a$ number of triangle areas respectively, and $\mathbf{K}_i$ is a block diagonal spring stiffness matrix.

\section{Experiments}   \label{section: experiments}
\begin{figure}[H]
\centering
\begin{minipage}[l]{0.10\textwidth}
\centering
\frame{\includegraphics[height=2cm,width=2cm]{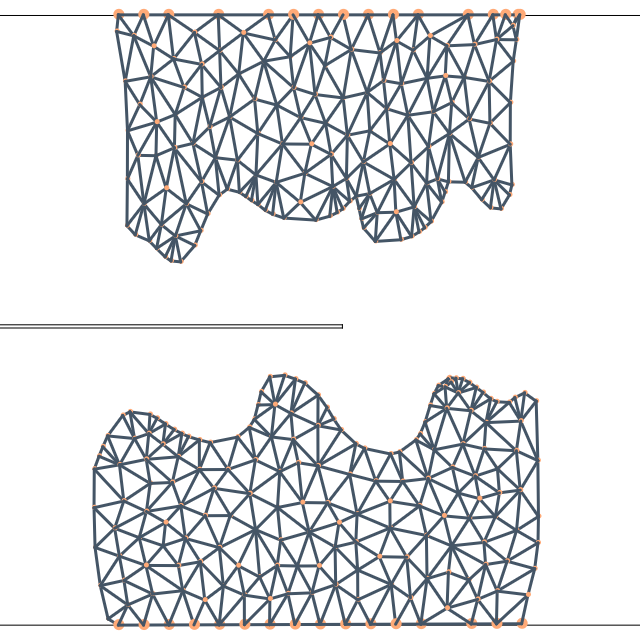}}
\vspace{-0.6cm}
\caption*{(a).Step=0}
\end{minipage}
\begin{minipage}[c]{0.10\textwidth}
\centering
\frame{\includegraphics[height=2cm,width=2cm]{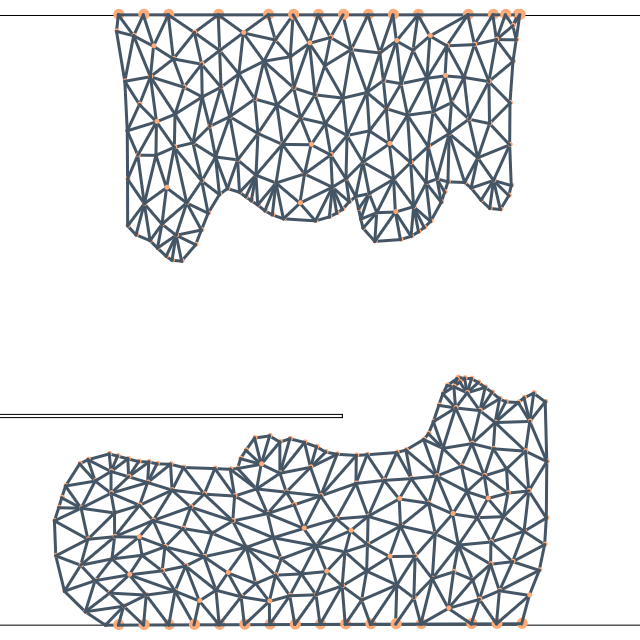}}
\vspace{-0.6cm}
\caption*{(b).Step=150}
\end{minipage}
\begin{minipage}[c]{0.10\textwidth}
\centering
\frame{\includegraphics[height=2cm,width=2cm]{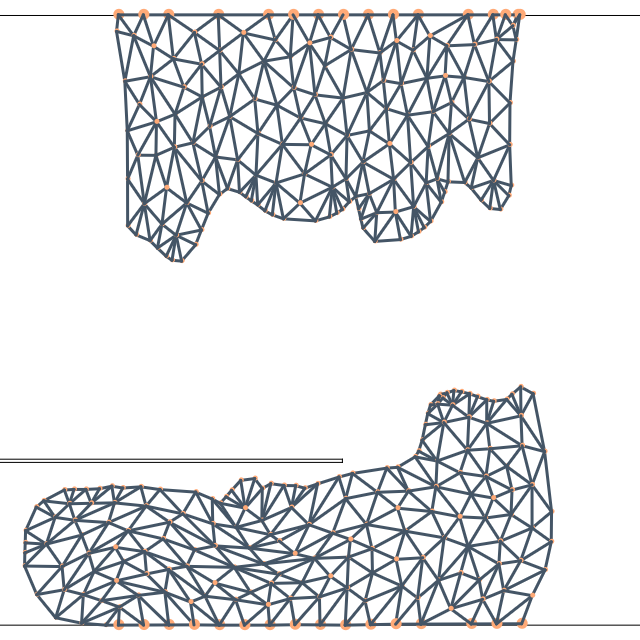}}
\vspace{-0.6cm}
\caption*{(c).Step=220}
\end{minipage}
\begin{minipage}[c]{0.10\textwidth}
\centering
\frame{\includegraphics[height=2cm,width=2cm]{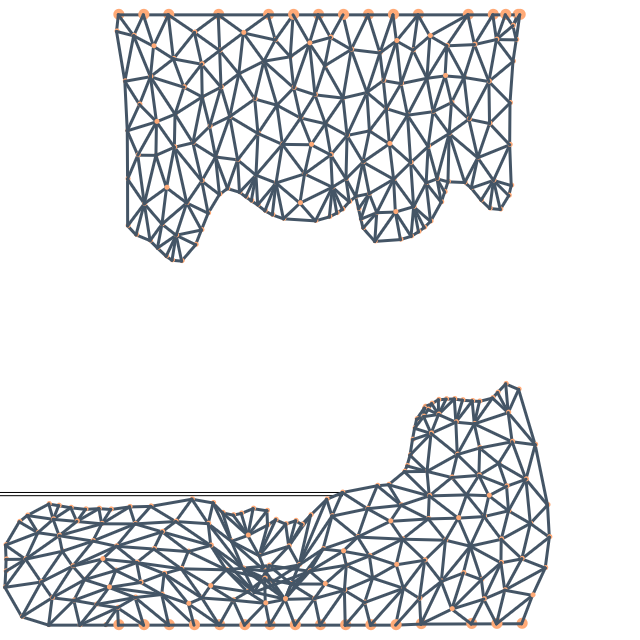}}
\vspace{-0.6cm}
\caption*{(d).Step=270}
\end{minipage}
\centering
\includegraphics[height=3.5cm, width=4.5cm]{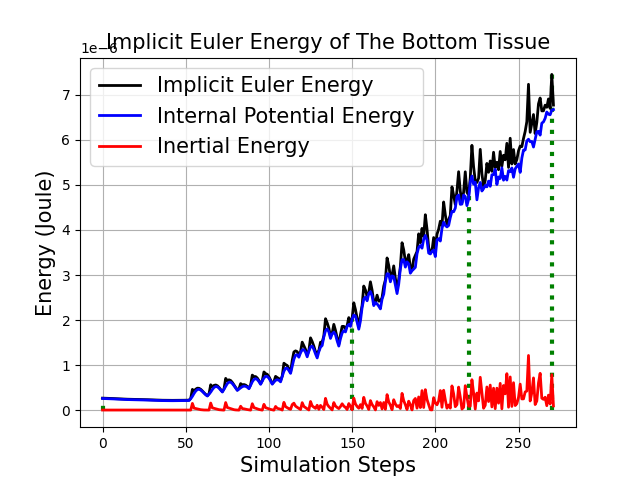}
\caption{The energy tracking when the tool is approaching the bottom tissue. The vertical green dash lines are corresponded to the upper figure simulation steps.}
\label{fig:approaching_bot}
\end{figure}

\begin{figure}[H]
\centering
\begin{minipage}[b]{0.45\textwidth}
\centering
\includegraphics[width=0.38\textwidth, height = 0.3\textwidth]{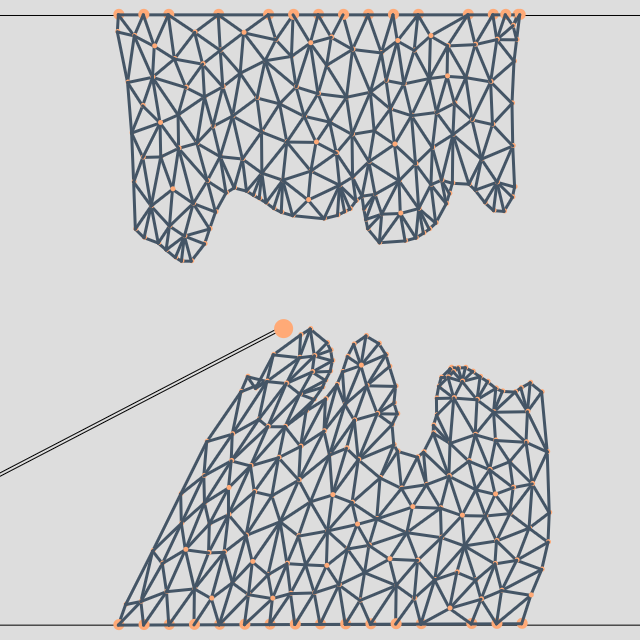} 
\includegraphics[width=0.38\textwidth,height = 0.3\textwidth]{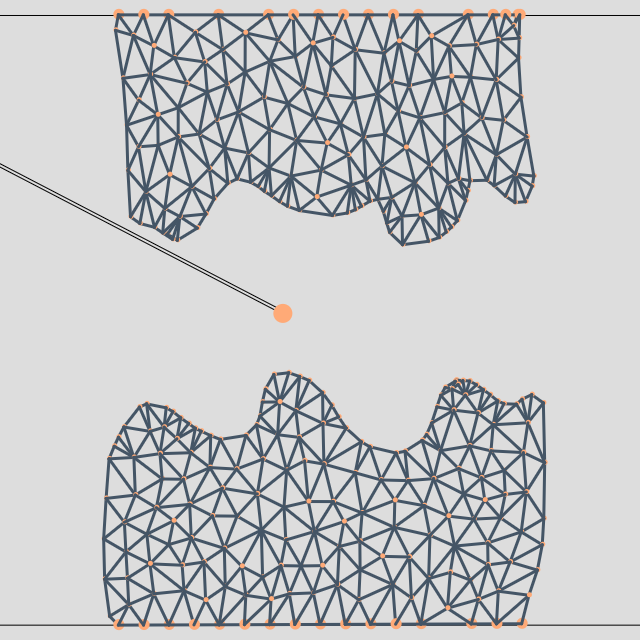}
\end{minipage}
\begin{minipage}[b]{0.45\textwidth}
\centering
\includegraphics[width=0.4\textwidth, height = 0.3\textwidth]{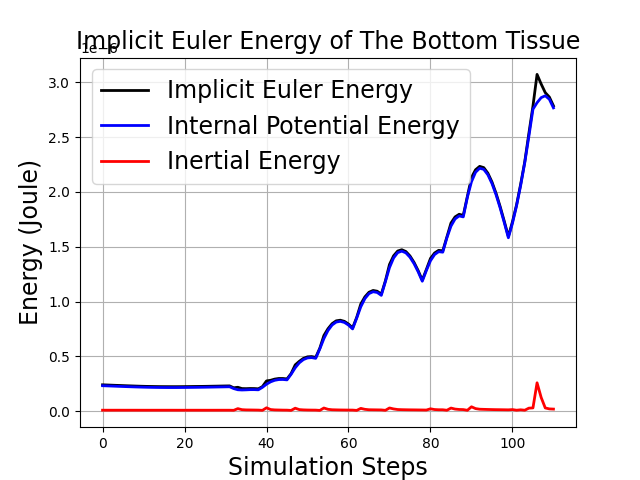} 
\includegraphics[width=0.4\textwidth,height = 0.3\textwidth]{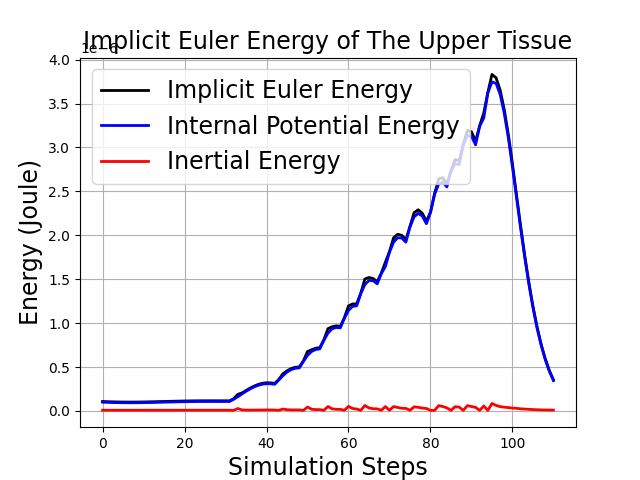}
\end{minipage}
\vspace{-0.2cm}
\caption{A demonstration of energy estimation for two different insertion angles. The top-left and top-right figures show the final tissue deformation when the manipulator reaches the target goal (yellow circle), respectively. The bottom-left and bottom-right figures show the implicit Euler energy in the simulation process, respectively. It is obviously to see the energy variation according to insertion angle.
}
\label{fig:diffangle_energy}
\end{figure}

In simulation, the boundary conditions are set as the upper and bottom horizontal lines shown in the right image in Fig \ref{fig:mesh_demo}. The velocity of the particle that crosses the horizontal lines are set as zero. The other simulation parameters (using metric system) are shown in Table \ref{tab:simu_para}. Two experiments have been done for different tasks.
\vspace{-0.2cm}
\begin{table}[H]
\centering
\begin{tabular}{ccccccccc}
\toprule  
$m_i$ & $k_{\text{spring}}$ & $k_{\text{area}}$ & $g$ & $\Delta t$ & Num$_{\text{Iter}}$ & r & $\alpha$\\
\midrule  
0.0001 & 0.15 & 1 & -9.8 & 0.01 & 30 & 0.25 & 0.025\\
\bottomrule 
\end{tabular}
\caption{Simulation parameters}
\label{tab:simu_para}
\end{table}
\vspace{-0.3cm}
\textbf{a.} The tool is approaching the bottom tissue and tracking of energy are shown in Fig. \ref{fig:approaching_bot}.
\vspace{-0.4cm}
\textbf{b.} We also track the energy variation of the tissues when the manipulator is inserted from two different angles which are aimed at the same target goal. As shown from Fig. \ref{fig:diffangle_energy}, it can be used to find optimal control and planning policies via energy based cost definition.

\vspace{-0.6cm}
\section{Conclusion} \label{sec:conclusion}
Our simulation framework can successfully simulate the 2D tissue deformations under the manipulation of the robotic manipulator. The energy computation is suitable for the control and planning applications. Our system can also be easily extended to 3D surgical environments by considering 3D mass-spring network and tetrahedral volume constraints.
\vspace{-0.5cm}

{\small
\bibliographystyle{./IEEEtran}
\bibliography{IEEEcitation}
}
\end{document}